\documentclass[letterpaper, 10 pt, final, conference]{ieeeconf}
\IEEEoverridecommandlockouts                              

\usepackage{amsmath,amsfonts}
\usepackage{algorithmic}
\usepackage{algorithm}
\usepackage{array}
\usepackage{textcomp}
\usepackage{url}
\usepackage{verbatim}
\usepackage{graphicx}
\usepackage{cite}
\usepackage{xcolor}
 
\title{\LARGE \bf
Hardware Implementation of a Zero-Prior-Knowledge Approach to Lifelong Learning in Kinematic Control of Tendon-Driven Quadrupeds
}

\author{Hesam Azadjou$^{*,1}$, Suraj Chakravarthi Raja$^{*,2}$, Ali Marjaninejad$^{1}$, Francisco J. Valero-Cuevas$^{1,2,3}$
\thanks{$^{*}$ First two authors contributed equally.}
\thanks{$^{1}$ Alfred E. Mann Department of Biomedical Engineering, University of Southern California, Los Angeles, CA 90089, USA.}
\thanks{$^{2}$ Ming Hsieh Department of Electrical and Computer Engineering, University of Southern California, Los Angeles, CA 90089, USA.}
\thanks{$^{3}$ Division of Biokinesiology and Physical Therapy, University of Southern California, Los Angeles, CA 90089, USA.}
\thanks{This project was partly supported by DARPA contract W911NF1820264, NIH grants R01-AR052345 and R01-AR050520, DoD-CDMRP grant MR150091, and NSF CRCNS Japan-US grant 2113096 awarded to F.J.V.-C.. This work was also partly funded by USC Research Enhancement Fellowships awarded to A.M. and S.C.R..}
\thanks{Contact: {\tt\small valero@usc.edu}}
}

\begin{document}

\maketitle
\thispagestyle{empty}
\pagestyle{empty}

\begin{abstract}
Like mammals, robots must rapidly learn to control their bodies and interact with their environment despite incomplete knowledge of their body structure and surroundings. They must also adapt to continuous changes in both. This work presents a bio-inspired learning algorithm, General-to-Particular (G2P), applied to a tendon-driven quadruped robotic system developed and fabricated in-house.
Our quadruped robot undergoes an initial five-minute phase of generalized motor babbling, followed by 15 refinement trials (each lasting 20 seconds) to achieve specific cyclical movements. This process mirrors the exploration-exploitation paradigm observed in mammals. With each refinement, the robot progressively improves upon its initial "good enough" solution.
Our results serve as a proof-of-concept, demonstrating the hardware-in-the-loop system's ability to learn the control of a tendon-driven quadruped with redundancies in just a few minutes to achieve functional and adaptive cyclical non-convex movements.
By advancing autonomous control in robotic locomotion, our approach paves the way for robots capable of dynamically adjusting to new environments, ensuring sustained adaptability and performance.
\end{abstract}

\begin{keywords}
Tendon/Wire Mechanism; Bioinspired Robot Learning; Continual Learning
\end{keywords}

\section{Introduction}
To operate in and move around the physical world for extended time periods, an agent must learn to control their body, understand the environment through the imperfect understanding of themselves, and update their understanding of both with time. The degree to which a learner, biological or synthetic, can imitate these abilities will be crucial in determining its capacity to learn, perform, and adapt in real-world applications with limited observability, incomplete or inaccurate priors, and uncertainties in interacting with the environment or other agents \cite{kudithipudi2022biological}. 
Evolutionary solutions for developing lifelong learning capabilities typically exhibit exceptional learning speed, efficiency, and adaptability, even without a detailed prior model of their body, the task, or the physics of the environment \cite{funk_testing_2007,Ye_lifelong_2023,azadjou2024play}. Prior works have demonstrated how vertebrates develop motor skills with minimal exposure to a task, adapt to new experiences, generalize basic principles across various tasks, and learn new tasks without overwriting existing ones \cite{chang_novo_2023,bartsch-jimenez_fine_2023,james_sim--real_2019, akkaya2019solving,bongard_resilient_2006,osa_hierarchical_2018,lowrey_reinforcement_2018}.

Vertebrates achieve this without explicit models of their body and environment physics ( or the parameters of these models such as maximal muscle forces, moment-arm values, etc.). Recent developments in robotics have started to investigate this approach to bio-inspired motor learning \cite{bonarini2006incremental,najjar2013self,urbina-melendez_brain-body-task_2024,azadjou2024play, bongard2006resilient, kwiatkowski2019task}. However, it continues to present a significant challenge when robots are expected to adapt to changes they encounter or learn in a physics-agnostic way, which opens new avenues for brain-body coevolution in robots.
We postulate that enabling robots to develop self-awareness of their continually changing bodies, and variations in environment dynamics, will enable them to quickly learn through exploration and adapt to changes in either their body or the environment with minimal loss in performance \cite{marjaninejad2019autonomous, kwiatkowski2019task}.

We chose tendon-driven systems as our study case for model agnostic autonomous learning of controls due to their bioinspired approach to movement and the inherent control challenges they pose. Unlike in torque-driven systems, tendon-driven systems can provide the system with great flexibility in actuator placement which can increase the movement efficiency \cite{marjaninejad2019should}. However, they are simultaneously over- and under-determined in nature which significantly constrains their feasible kinematic state space. This makes the control problem more challenging, as there is no one-to-one relationship between the degrees of freedom (DoFs) and actuators \cite{valero2016fundamentals}.

In this study, we have implemented and expanded General-to-Particular (G2P) algorithm \cite{marjaninejad2019autonomous} to learn the kinematic control of a tendon-driven quadruped robot where each limb has two DoFs controlled by three tendons. Our results indicate that the quadruped robot can learn to follow non-convex non-differentiable cyclical movements within five minutes of initial babbling, demonstrating control proficiency coupled with rapid adaptation. Continuous refinements through sensory feedback and a few-shot learning pipeline enable the robot to learn to interact with their body and environment with little-to-zero prior knowledge of either. This approach serves as a proof-of-principal demonstration for the potential of lifelong learning in advancing the state of autonomous movement control for real-world hardware implementation of bio-inspired robots. By imitating vertebrate-like learning processes, our approach paves the way for robots that can dynamically and autonomously adjust to new environments and tasks, ensuring persistent and adaptable performance. At the same time, it can be a test bed to better study of mechanics, kinematics, and dynamics of biological systems.

\section{Methods}
\subsection{Quadruped design in Hardware} \label{sec:quadruped_design_hard}
\begin{figure}
\centering
\includegraphics[width=\linewidth, keepaspectratio]{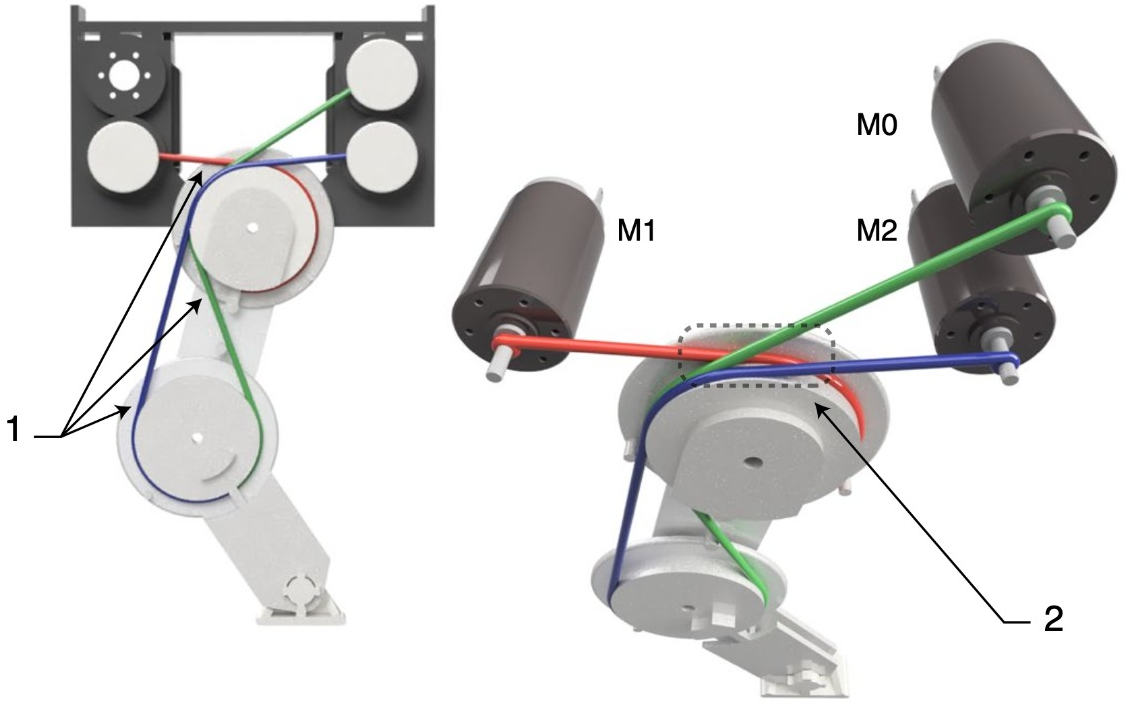}
\caption{\textbf{Quadruped Leg Structure:} Three tendons (label \textbf{1}) are controlled by motors M0, M1, and M2. The mono-articular tendon connected to motor M1 exclusively drives the proximal joint in an counter-clockwise (flexion) direction. However the tendons connected to motors M0 and M2 are multi-articular. Motor M0 drives the proximal joint clockwise (extension) and the distal joint counter-clockwise (flexion). Motor M2 only performs extension on both joints. Tendon channel (see label \textbf{2}) keep the tendons from falling off the joints. The mechanical design of the legs (and this figure) was obtained from a previous study (Marjaninejad et al.\cite{marjaninejad2019autonomous}).}
\label{fig:tendon_routing_leg}
\end{figure}

The tendon routing of each quadruped's the planar legs is illustrated in figure \ref{fig:tendon_routing_leg}. The orientation of the hind legs were mirrored with respect to fore legs.
This tendon-driven quadruped featured light-weight designs conducive to additive manufacturing (3D printing) such that the knee joints bent inwards.
Motor and leg mounts were designed to be modular and easily interchangeable. This was essential since part failures happened when the mechanical stresses exceeding the parts’ tensile strength and/or when the temperature of the motors exceeded the glass transition temperature of polylactic acid (PLA) material causing the 3D printed motor mounts to warp under load.
The hip and shoulder structures were also mirror symmetric, possessing a tetrahedral design to maximize robustness to mechanical stresses while reducing weight. 
Hard stops at the hip and knee restricted joint angles to ensure that the feasible force spaces of the end effectors never collapsed \cite{valero2016fundamentals}. In turn, this ensured that the quadruped would never get trapped in singularities in the cost landscape.

The electromechanical architecture for sensing and tendon actuation was based on a previous study involving locomotion and movement in tendon-driven systems \cite{jalaleddini2017neuromorphic, marjaninejad2019autonomous, chakravarthirajaNovelPassiveImplantable2023}.
Each of the three inelastic nylon tendons of a limb was actuated by a Faulhaber\textsuperscript{\textregistered} 2342-S024-CR gear-free brushed DC motor. The gear-free motor together with a small coil of tendon on each motor shaft ensured backdrivability. The motors were driven by Western Servo Design LDU-S1 linear current amplifiers. Voltage commands to these amplifiers were used to modulate the DC current through and hence the torque applied by each motor. The hip and knee joint angles for each limb were sensed by CUI\textsuperscript{\textregistered} AMT103-V incremental encoders \cite{cui_amt103-v}.

We used a custom \textit{Marionette} toolchain for tendon-driven robotics was used for low-level robot operation and remote control \cite{ChakravarthiRaja_thesis2023}. One of its components--- \textit{QuickDAQ}, a software data acquisition library that interfaced with National Instruments\textsuperscript{\textregistered} (NI) DAQmx data drivers---was used to read encoder angles and motor voltage commands from a PXI\textsuperscript{\textregistered} data acquisition PC computer. The toolchain was also used for closed-loop PI control of tendon tension. In-turn, G2P-based machine learning (ML) pipeline (see section \ref{sec:G2P_learning_pipeline}) operating on a separate ML computer received encoder angle data from, and returned motor reference commands to the PXI PC. The Sub-millisecond near-real-time communication between the two computers was achieved over a peer-to-peer Ethernet connection using \emph{RT-Bridge}, another component of Marionette which enables remote control and telemetry using the ZeroMQ message-passing library \cite{hintjens2013zeromq, lauener2017design}.

\begin{figure*}
\centering
\includegraphics[width=0.99\linewidth, keepaspectratio]{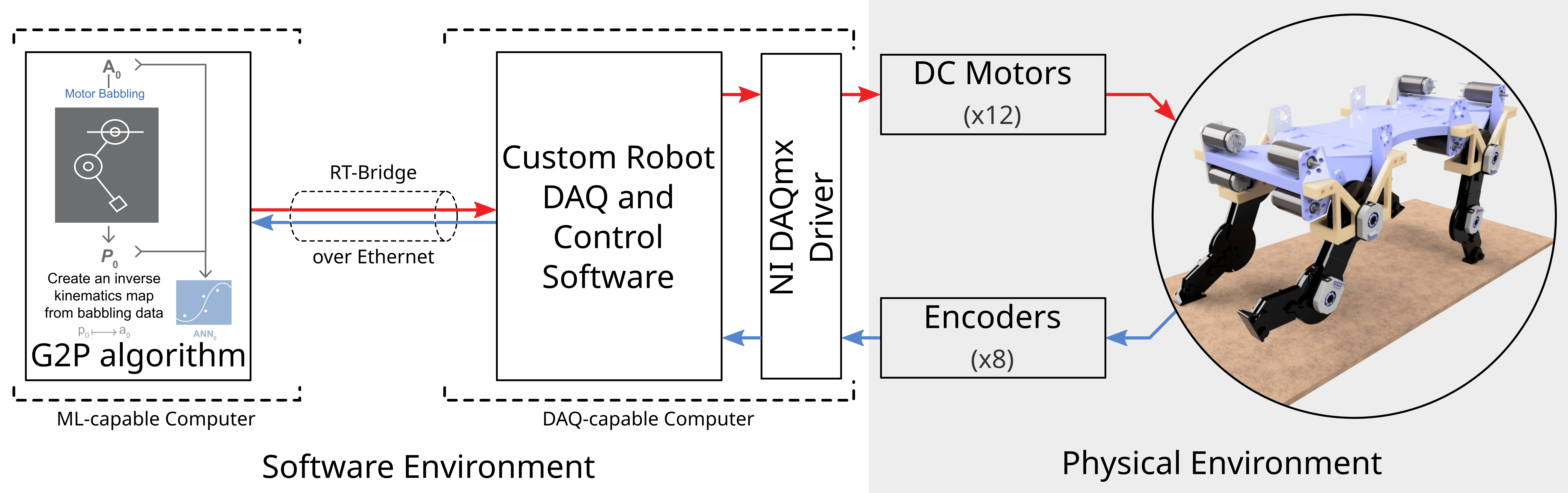}
\caption{\textbf{System Overview:} The physical implementation of the robot features software and hardware sections. Our modular tendon-driven robot, featuring DC motors and encoders, was controlled by a custom DAQ and control program on a DAQ-capable computer running the NI-DAQmx driver. The G2P learning pipeline---running on a machine learning-capable computer---used RT-Bridge over Ethernet to send commands and receive sensory data from the DAQ program in near-real-time with sub-millisecond round-trip latency.}
\label{fig:physical_implementation}
\end{figure*}

The architecture of our physical implementation, featuring both the physical and software environments is illustrated in figure \ref{fig:physical_implementation}.
 
\subsection{Task and Test Case}
\subsubsection{Task}
We have tested all our cases with continuous cyclical movements.
During this task, both proximal and distal joints follow sinusoidal trajectories with a $\pi/2$ phase difference. Additionally, diagonal limbs are synchronized together (similar to trotting). This task consists of 20 seconds of movement with 3.6 second cycles.

\subsubsection{Test Case}
We have tested our quadruped while it was suspended in air, so its limbs did not contact the floor. Therefore, the system only needed to learn how to control the dynamics of its own limbs.

\subsection{Dynamic Control of Tendon-Driven Legs}
Our system controls four tendon-driven legs, enabling a quadruped to walk using four DC motors. Each motor pulls a tendon, adjusting the corresponding leg's position and velocity. The control strategy dynamically activates the motors based on a model that classifies movement patterns and their intensities at different time points.

Mathematically speaking, each leg \( L_i \) may be modeled as a second-order system:
\begin{equation}
\label{eq:leg_dynamics}
\ddot{q} = -I(q)^{-1}C(q, \dot{q}) + B\dot{q} + I(q)^{-1}T
\end{equation}
where \( q \) is the joint angle, \( I \) is the inertia matrix, \( C(q, \dot{q}) \) is the Coriolis and centripetal forces matrix, \( B \) is the damping coefficients matrix, and \( T(t) \) is the torque at the joints.

The forces exerted by the musculotendons (represented here as cables pulled by motors) are subsequently linked to the vector of applied joint torques as:
\begin{equation}
T(t) = M(q)F_0a
\end{equation}
where \( M(q) \) is the moment arm matrix, \( F_0 \) is a diagonal matrix containing the maximum force that each actuator can exert, and \( a \) is the activation sequence sent to the actuator. The system aims to achieve the desired leg positions \( x_{i, \text{des}}(t) \) and apply forces corresponding to the desired activations \( a \).

We trained an artificial neural network (ANN) to be an inverse map connecting the resulting kinematics with each random activation that produced them. The inverse map aims to find the actuation values vector \( a \) for any given set of desired kinematics \( q, \dot{q}, \ddot{q} \) without using any implicit model and only from the babbling and task-specific data. The trained ANN for mapping used in the lower-level control of this study is described in the equation:
\begin{equation}
\label{eq:ann_mapping}
a = ANN(q, \dot{q}, \ddot{q})
\end{equation}

The control system adjusts the voltage applied to each motor driver (which in turn modulates the motor toque) using the activations predicted by the ANN, ensuring the legs achieve the desired movements and intensities in real-time.

\subsection{Learning Pipeline (G2P Algorithm)} \label{sec:G2P_learning_pipeline}
To find a mapping between desired kinematics and muscle activations in our tendon-driven system, we use the G2P algorithm~\cite{marjaninejad2019autonomous}, which consists of two main parts:

\subsubsection{Motor Babbling}
During this phase, muscles are randomly activated (uniformly distributed from 0 to 100\% activation), and the resulting sensory input information (tactile sensory information, such as endpoint force values, and kinematics, including joint angles, angular velocities, and angular accelerations) are collected. These sensory inputs and activations are then used to train an ANN as inputs and desired outputs, respectively.

The resulting ANN is used to predict muscle activations required to perform the desired tasks during the refinement phase. For all cases in this study, we performed the babbling phase for 60 seconds.

\subsubsection{Refinements}
During this phase, the kinematics of the desired task (either cyclical or point-to-point) are sent to the ANN to estimate the required muscle activations. These activations are then used to perform the movement. For tactile sensory input, we feed the previous time step's sensory data (except for the first simulation step, where we feed 0). The resulting task-specific sensory information and muscle activations are concatenated with the data available so far and used to re-tune the ANN. Note that motor babbling provides sparse sampling within a vast volume of sensory information, while refinements enable sampling more specific to the sensory space of a desired task.

\subsection{Scaling Sensory Data}
Scaling and normalizing input data can enhance learning speed and improve the data efficiency of a machine learning algorithm, especially when inputs have different units and ranges. We scaled the input data by dividing them by their expected variance. These scaling factors were calculated by running a 60-second babbling (done only once for the entire curriculum of tasks and test cases) with the quadruped in the air (no load). This approach helps in faster convergence of the ANN training processes.

\subsection{ANN Architecture}
We use a single ANN for each limb (a total of four ANNs, each having identical network architecture) that maps all the sensory input of that limb to the predicted muscle activation values (3 muscles). The ANNs start with babbling (in air), and once trained, they concatenate kinematic data (position, velocity, and acceleration for each of the two joint) from new babbling or refinements and re-tune their weights using this cumulative data set, warm-starting with the weights from the last case. The error used to train the ANNs is the mean square error (MSE) over all muscle activations. We use multilayer perceptron (MLP) ANNs with one hidden layer (24 neurons for the single ANN and 6 neurons for each ANN in the multiple ANN case), linear activation functions (which performed better than sigmoid functions), and the ADAM optimizer~\cite{kingma2014adam} in the Keras API from the TensorFlow library.

\subsection{Position Error Feedback}
Similar to~\cite{marjaninejad2019simple}, we implemented corrective position error feedback on the joints' position error. To measure system performance on the desired movements, we calculate the root mean square error (RMSE) over the joint angles (across all limbs) for the last half of the data (to avoid effects of transient initial conditions) and report it. RMSE is used instead of MSE because it preserves the units of the inputs (radians).

\section{Results and discussion}
In this study, we implemented a method for a quadruped robot to learn cyclical movements in air using a structured approach based on motor babbling and ANN-based inverse mapping. The following sections detail the steps taken and the outcomes observed.

\subsection{Initial Motor Babbling}
Each run starts with a five-minute motor babbling about initiating a mapping between the kinematics and motor activation sequences pairs for the quadruped. During this phase, a pseudo-random control sequence was fed to the motors of each limb, generating a 3D time-varying vector of current changes. The kinematic data (joint angles, angular velocities, and angular accelerations) were recorded using encoders at each joint. These data formed the basis for training an initial artificial neural network (ANN) to create an inverse map from 6D kinematics to a 3D control sequence.

\subsection{Movement Parameterization and Control Sequence Generation}
A closed orbit in 2D joint-angle space, defined by a 'feature vector' of 10 evenly distributed points, parameterized the desired cyclical movement (see \cite{marjaninejad2019autonomous}). For a cycle duration of approximately 1 second, this orbit defined the 6D limb kinematics for each of the robot's four limbs. Using the initial inverse map, we produced a control sequence, which was applied to the quadruped's limbs, generating $\approx$5 movement cycles.

\subsection{Performance Evaluation}
\begin{figure}
    \centering
    \includegraphics[width=0.99\linewidth, keepaspectratio]{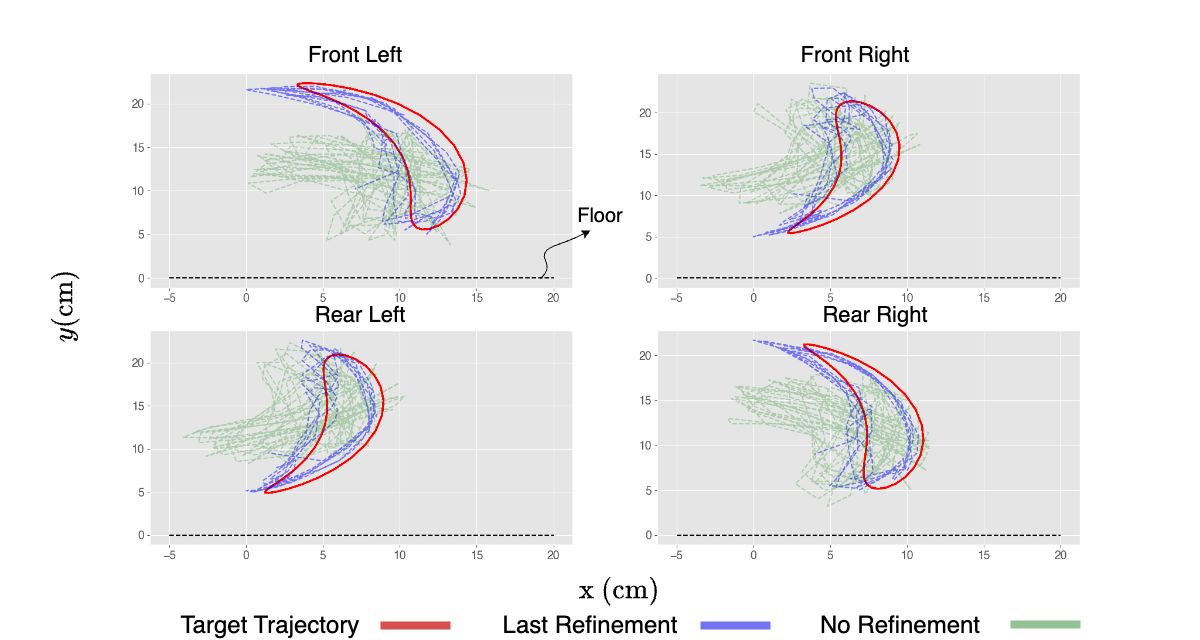}
    \caption{\textbf{Endpoint trajectories of the quadruped Suspended in air:} The green trajectories represent the performance immediately after babbling, with no refinement. The solid red trajectory indicates the target path the quadruped was supposed to follow. The blue trajectory illustrates the performance after the final refinement. This figure demonstrates the quadruped's progress following the target trajectory through successive refinements. }
    \label{fig:endpoints}
\end{figure}

\begin{figure}
    \centering
    \includegraphics[width=0.99\linewidth, keepaspectratio]{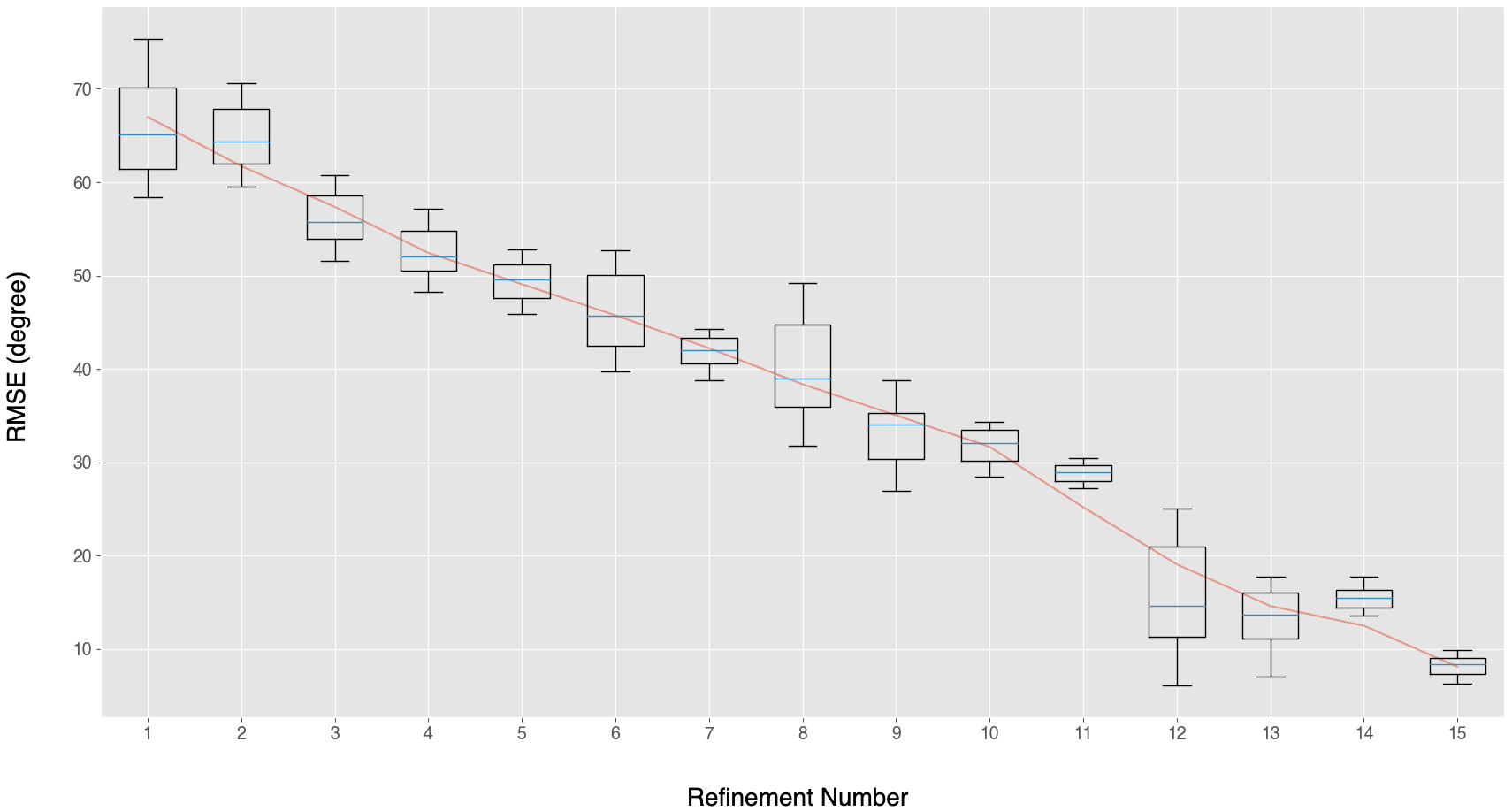}
    \caption{\textbf{Refinements in Air:} 15 refinements of a single over five minutes of continuous operation were performed in the air following motor babbling. The RMSE (Root Mean Square Error) of the resulting trajectories was plotted for the two joints of each of the four legs (eight values for each boxplot). Initially, the RMSE was $65.4^\circ$, which was reduced to $9.3^\circ$ by the end of the refinements.}
    \label{fig:box}
\end{figure}

After each control sequence application, the resulting kinematics were appended to the dataset, which included initial motor babbling and previous attempts. This aggregated data was used to refine the ANN gradually. Our quadruped learned to follow the particular cyclical movement (see Fig. \ref{fig:endpoints}) within 15 refinements (each lasting only 20 seconds) following an initial five-minute period of generalized motor babbling, emulating the exploration-exploitation paradigm seen in mammals. The system demonstrated adaptive improvement, mirroring trial-to-trial experiential adaptation observed in biological motor learning (Fig. \ref{fig:box}).

\begin{figure}
    \centering
    \includegraphics[width=0.99\linewidth, keepaspectratio]{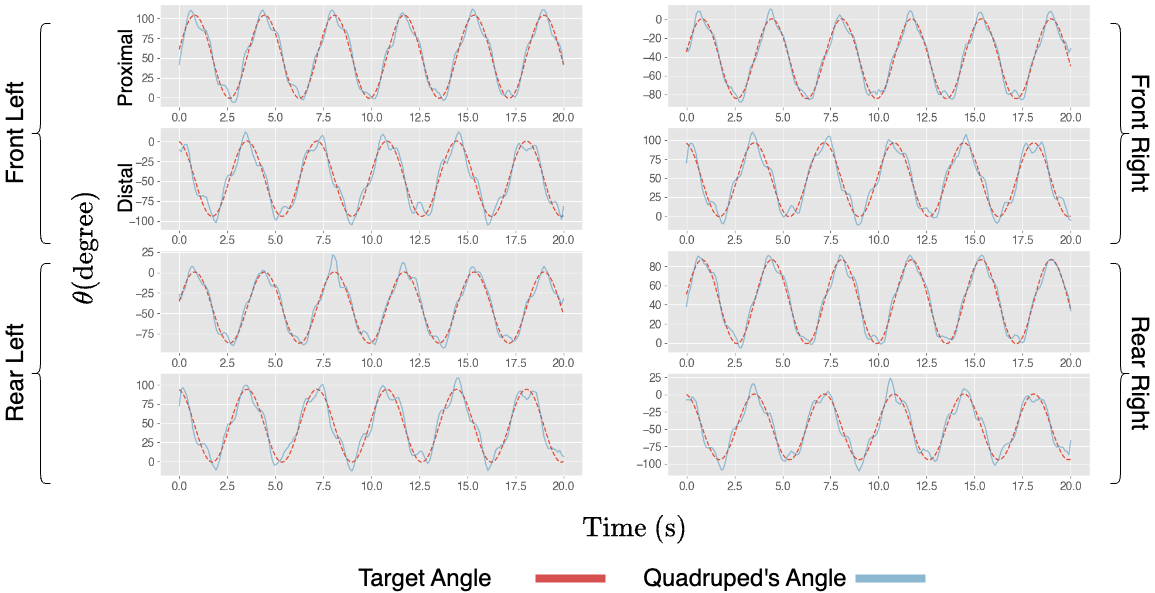}
    \caption{\textbf{Joint Angles:} The real joint angles were plotted vs the target trajectories after the 15\textsuperscript{th} refinement showing the quality of the performance of the task by the legs for a cyclical motion in air (without load) with zero knowledge of the body and the environment before the motor babbling.}
    \label{fig:traj}
\end{figure}

After 15 refinements within the same run, the quadruped was able to follow the target angle trajectories (Fig. \ref{fig:traj}) both in the distal and proximal joints. The endpoint locations resulting from this reproduction could follow the desired trajectory (Fig. \ref{fig:endpoints}), a non-convex curve, serving as a proof-of-principle for the G2P algorithm in hardware. Through each refinement step, the robot continually improves on its already "good enough" solution to progressively reduce the RMSE. After each refinement, the previous refinement iteration's ANN is transferred and further trained on the new refinement data. This process fine-tunes the mapping parameters, enabling continuous improvement and lifelong learning. The ANN adapts and evolves by progressively integrating new data, ensuring more accurate and effective performance over time. When the robot runs for longer durations, this approach of repetitive refinement could be leveraged to enable the robot to adapt to wear-and-tear in its body and changes in the environment, achieving functional and adaptive cyclical non-convex movements without prior knowledge of the environment or its own body.

These results are promising, demonstrating the hardware-in-the-loop system's ability to learn the control of a tendon-driven quadruped with redundancies in just a few minutes. The system's ability to continually learn and adapt suggests potential lifelong learning capabilities, advancing the state-of-the-art in autonomous control of robotic locomotion by offering a bio-inspired approach to realizing robots that can dynamically and autonomously adjust to new environments, ensuring persistent and adaptable performance.

\subsection{Conclusion}
The quadruped robot successfully learned non-convex non-differentiable cyclical movements in air, leveraging just five minutes of motor babbling coupled with an ANN-based learned inverse map. The refinement steps in G2P enable online learning without prior knowledge. This allowed our quadrupeds to display continual improvement in performance with each refinement. Our results show that the hardware-in-the-loop system’s ability to learn the control of a tendon-driven quadruped with redundancies in just ten minutes (five minutes each of babbling and refinements) to achieve functional and adaptive cyclical non-convex movements. This implementation underscores the efficacy of structured learning approaches for robotic motor control.

\subsection{Limitation and Future Work}
Artificial intelligence is a rapidly growing field that has been beneficial in various domains of engineering \cite{shafiee2024effect,bideh2018security,baraeinejad2022design}. In this example, we demonstrate how using ANN architectures enables a quadruped robot to learn limb control from scratch. The ability to adapt to varying conditions is a crucial step for robots to successfully perform locomotion tasks. \cite{ames2017hybrid, vukobratovic2004zero, holmes2006dynamics, kuo2007six, iqbal2014bifurcations}. But, while our results demonstrate the ability of G2P learning pipeline to achieve kinematic control of our quadrupedal robot, our approach has not been tested in the context of the dynamics associated with either loading the robot or subjecting it to different operational environments.

Another factor that is key to locomotion is the ability to balance the body on uneven terrains \cite{kuo1999stabilization, harada2006dynamics}. However, our robot limbs are entirely planar and unable to generate three dimensional force spaces at the end-effectors---an important property to achieve effective balance control. The increase in the number of DoFs to achieve this would in-turn necessitate more tendons. This would result in an increase in the dimensionality of the control/activation space which the learning pipeline would need to explore. Unless handled carefully, this would greatly increase the babbling time of our robots.

To address these limitations, we have a series of improvements on which we are currently working. As a first step, we are currently using a gantry-based system to test our robot and G2P learning algorithm's ability to perform locomotion at different simulated boy weights. Then, we will test the ability of G2P learning methods to learn to switch between multiple tasks such as standing from prone position, walking in a straight line, and turning. In the mean time, we are working on novel edge accelerator designs for the real-time acceleration of learning pipelines for larger robotic systems, and bio-inspired distributed neural control architectures to reduce the dimensionality of the control space \cite{ChakravarthiRaja_thesis2023, raphael2010spinal, niyo2023alpha}. Finally, we intend replace the planar limbs of our quadruped with limbs capable of movement and force production in 3D space.

\section*{ACKNOWLEDGMENT}
The authors thank Dr. Darío Urbina-Melendez for leading the design and construction of the mechanical body of the quadruped. Likewise, we would also like to thank Mr. Timothy Fanelle for assisting us during the early phases of the study with both debugging the control architecture of the system and running initial pilot experiments.

\bibliographystyle{IEEEtran}
\bibliography{root}

\begin{thebibliography}{10}
\providecommand{\url}[1]{#1}
\csname url@rmstyle\endcsname
\providecommand{\newblock}{\relax}
\providecommand{\bibinfo}[2]{#2}
\providecommand\BIBentrySTDinterwordspacing{\spaceskip=0pt\relax}
\providecommand\BIBentryALTinterwordstretchfactor{4}
\providecommand\BIBentryALTinterwordspacing{\spaceskip=\fontdimen2\font plus
\BIBentryALTinterwordstretchfactor\fontdimen3\font minus \fontdimen4\font\relax}
\providecommand\BIBforeignlanguage[2]{{%
\expandafter\ifx\csname l@#1\endcsname\relax
\typeout{** WARNING: IEEEtran.bst: No hyphenation pattern has been}%
\typeout{** loaded for the language `#1'. Using the pattern for}%
\typeout{** the default language instead.}%
\else
\language=\csname l@#1\endcsname
\fi
#2}}

\bibitem{kudithipudi2022biological}
D.~Kudithipudi, M.~Aguilar-Simon, J.~Babb, M.~Bazhenov, D.~Blackiston, J.~Bongard, A.~P. Brna, S.~Chakravarthi~Raja, N.~Cheney, J.~Clune, \emph{et~al.}, ``Biological underpinnings for lifelong learning machines,'' \emph{Nature Machine Intelligence}, vol.~4, no.~3, pp. 196--210, 2022.

\bibitem{funk_testing_2007}
\BIBentryALTinterwordspacing
M.~S. Funk and P.~Tosto, ``Testing {Skills} in {Vertebrates},'' \emph{The American Biology Teacher}, vol.~69, no.~2, pp. 85--91, Feb. 2007, publisher: National Association of Biology Teachers. [Online]. Available: \url{https://bioone.org/journals/the-american-biology-teacher/volume-69/issue-2/0002-7685_2007_69_85_TSIV_2.0.CO_2/Testing-Skills-in-Vertebrates/10.1662/0002-7685(2007)69[85:TSIV]2.0.CO;2.full}
\BIBentrySTDinterwordspacing

\bibitem{Ye_lifelong_2023}
\BIBentryALTinterwordspacing
F.~Ye and A.~G. Bors, ``Lifelong generative adversarial autoencoder,'' \emph{IEEE Transactions on Neural Networks and Learning Systems}, p. 1–15, 2023. [Online]. Available: \url{http://dx.doi.org/10.1109/TNNLS.2023.3281091}
\BIBentrySTDinterwordspacing

\bibitem{azadjou2024play}
H.~Azadjou, A.~Marjaninejad, and F.~Valero-Cuevas, ``Play it by ear: A perceptual algorithm for autonomous melodious piano playing with a bio-inspired robotic hand,'' \emph{bioRxiv}, pp. 2024--06, 2024.

\bibitem{chang_novo_2023}
\BIBentryALTinterwordspacing
J.~C. Chang, M.~G. Perich, L.~E. Miller, J.~A. Gallego, and C.~Clopath, ``\BIBforeignlanguage{en}{De novo motor learning creates structure in neural activity space that shapes adaptation},'' May 2023, pages: 2023.05.23.541925 Section: New Results. [Online]. Available: \url{https://www.biorxiv.org/content/10.1101/2023.05.23.541925v2}
\BIBentrySTDinterwordspacing

\bibitem{bartsch-jimenez_fine_2023}
\BIBentryALTinterwordspacing
A.~Bartsch-Jimenez, M.~Błażkiewicz, H.~Azadjou, R.~Novotny, and F.~J. Valero-Cuevas, ``Fine synergies” describe motor adaptation in people with drop foot in a way that supplements traditional “coarse synergies,'' \emph{Frontiers in sports and active living}, vol.~5, p. 1080170, 2023, publisher: Frontiers. [Online]. Available: \url{https://www.frontiersin.org/articles/10.3389/fspor.2023.1080170/full}
\BIBentrySTDinterwordspacing

\bibitem{james_sim--real_2019}
\BIBentryALTinterwordspacing
S.~James, P.~Wohlhart, M.~Kalakrishnan, D.~Kalashnikov, A.~Irpan, J.~Ibarz, S.~Levine, R.~Hadsell, and K.~Bousmalis, ``Sim-to-{Real} via {Sim}-to-{Sim}: {Data}-efficient {Robotic} {Grasping} via {Randomized}-to-{Canonical} {Adaptation} {Networks},'' July 2019, arXiv:1812.07252 [cs]. [Online]. Available: \url{http://arxiv.org/abs/1812.07252}
\BIBentrySTDinterwordspacing

\bibitem{akkaya2019solving}
I.~Akkaya, M.~Andrychowicz, M.~Chociej, M.~Litwin, B.~McGrew, A.~Petron, A.~Paino, M.~Plappert, G.~Powell, R.~Ribas, \emph{et~al.}, ``Solving rubik's cube with a robot hand,'' \emph{arXiv preprint arXiv:1910.07113}, 2019.

\bibitem{bongard_resilient_2006}
\BIBentryALTinterwordspacing
J.~Bongard, V.~Zykov, and H.~Lipson, ``Resilient {Machines} {Through} {Continuous} {Self}-{Modeling},'' \emph{Science}, vol. 314, no. 5802, pp. 1118--1121, Nov. 2006, publisher: American Association for the Advancement of Science. [Online]. Available: \url{https://www.science.org/doi/10.1126/science.1133687}
\BIBentrySTDinterwordspacing

\bibitem{osa_hierarchical_2018}
\BIBentryALTinterwordspacing
T.~Osa, J.~Peters, and G.~Neumann, ``\BIBforeignlanguage{en}{Hierarchical reinforcement learning of multiple grasping strategies with human instructions},'' \emph{\BIBforeignlanguage{en}{Advanced Robotics}}, vol.~32, no.~18, pp. 955--968, Sept. 2018. [Online]. Available: \url{https://www.tandfonline.com/doi/full/10.1080/01691864.2018.1509018}
\BIBentrySTDinterwordspacing

\bibitem{lowrey_reinforcement_2018}
\BIBentryALTinterwordspacing
K.~Lowrey, S.~Kolev, J.~Dao, A.~Rajeswaran, and E.~Todorov, ``\BIBforeignlanguage{en}{Reinforcement learning for non-prehensile manipulation: {Transfer} from simulation to physical system},'' in \emph{\BIBforeignlanguage{en}{2018 {IEEE} {International} {Conference} on {Simulation}, {Modeling}, and {Programming} for {Autonomous} {Robots} ({SIMPAR})}}.\hskip 1em plus 0.5em minus 0.4em\relax Brisbane, QLD: IEEE, May 2018, pp. 35--42. [Online]. Available: \url{https://ieeexplore.ieee.org/document/8376268/}
\BIBentrySTDinterwordspacing

\bibitem{bonarini2006incremental}
A.~Bonarini, A.~Lazaric, and M.~Restelli, ``Incremental skill acquisition for self-motivated learning animats,'' in \emph{International Conference on Simulation of Adaptive Behavior}.\hskip 1em plus 0.5em minus 0.4em\relax Springer, 2006, pp. 357--368.

\bibitem{najjar2013self}
T.~Najjar and O.~Hasegawa, ``Self-organizing incremental neural network (soinn) as a mechanism for motor babbling and sensory-motor learning in developmental robotics,'' in \emph{International Work-Conference on Artificial Neural Networks}.\hskip 1em plus 0.5em minus 0.4em\relax Springer, 2013, pp. 321--330.

\bibitem{urbina-melendez_brain-body-task_2024}
\BIBentryALTinterwordspacing
D.~Urbina-Meléndez, H.~Azadjou, and F.~J. Valero-Cuevas, ``Brain-{Body}-{Task} {Co}-{Adaptation} can {Improve} {Autonomous} {Learning} and {Speed} of {Bipedal} {Walking},'' Feb. 2024, arXiv:2402.02387 [cs]. [Online]. Available: \url{http://arxiv.org/abs/2402.02387}
\BIBentrySTDinterwordspacing

\bibitem{bongard2006resilient}
J.~Bongard, V.~Zykov, and H.~Lipson, ``Resilient machines through continuous self-modeling,'' \emph{Science}, vol. 314, no. 5802, pp. 1118--1121, 2006.

\bibitem{kwiatkowski2019task}
R.~Kwiatkowski and H.~Lipson, ``Task-agnostic self-modeling machines,'' \emph{Science Robotics}, vol.~4, no.~26, p. eaau9354, 2019.

\bibitem{marjaninejad2019autonomous}
A.~Marjaninejad, D.~Urbina-Mel{\'e}ndez, B.~A. Cohn, and F.~J. Valero-Cuevas, ``Autonomous functional movements in a tendon-driven limb via limited experience,'' \emph{Nature machine intelligence}, vol.~1, no.~3, p. 144, 2019.

\bibitem{marjaninejad2019should}
A.~Marjaninejad and F.~J. Valero-Cuevas, ``Should anthropomorphic systems be “redundant”?'' in \emph{Biomechanics of Anthropomorphic Systems}.\hskip 1em plus 0.5em minus 0.4em\relax Springer, 2019, pp. 7--34.

\bibitem{valero2016fundamentals}
F.~J. Valero-Cuevas, \emph{Fundamentals of neuromechanics}.\hskip 1em plus 0.5em minus 0.4em\relax Springer, 2016, vol.~8.

\bibitem{jalaleddini2017neuromorphic}
K.~Jalaleddini, C.~M. Niu, S.~C. Raja, W.~J. Sohn, G.~E. Loeb, T.~D. Sanger, and F.~J. Valero-Cuevas, ``Neuromorphic meets neuromechanics, part ii: the role of fusimotor drive,'' \emph{Journal of neural engineering}, vol.~14, no.~2, p. 025002, 2017.

\bibitem{chakravarthirajaNovelPassiveImplantable2023}
S.~Chakravarthi~Raja, W.~S. You, K.~Jalaleddini, J.~C. Casebier, N.~R. {Lightdale-Miric}, V.~R. Hentz, F.~J. {Valero-Cuevas}, and R.~Balasubramanian, ``A {{Novel Passive Implantable Differential Mechanism}} to {{Restore Individuated Finger Flexion}} during {{Grasping}} following {{Tendon Transfer Surgery}}: {{A Pilot Study}},'' \emph{Applied Sciences}, vol.~13, no.~9, p. 5804, Jan. 2023.

\bibitem{cui_amt103-v}
\BIBentryALTinterwordspacing
{AMT}103-v. [Online]. Available: \url{https://www.digikey.com/en/products/detail/cui-devices/AMT103-V/827016}
\BIBentrySTDinterwordspacing

\bibitem{ChakravarthiRaja_thesis2023}
\BIBentryALTinterwordspacing
S.~Chakravarthi~Raja, ``Marionette: A neurorobotics toolchain for the neuromorphic control of tendon-driven systems,'' Ph.D. dissertation, University of Southern California, Los Angeles, California, United States, Sept. 2023. [Online]. Available: \url{https://digitallibrary.usc.edu/asset-management/2A3BF1MH6TP1G}
\BIBentrySTDinterwordspacing

\bibitem{hintjens2013zeromq}
P.~Hintjens, \emph{ZeroMQ: messaging for many applications}.\hskip 1em plus 0.5em minus 0.4em\relax " O'Reilly Media, Inc.", 2013.

\bibitem{lauener2017design}
J.~Lauener, W.~Sliwinski, and G.~CERN, ``How to design \& implement a modern communication middleware based on zeromq,'' in \emph{Proc of ICALEPCS}, vol.~17, 2017, pp. 45--51.

\bibitem{kingma2014adam}
D.~P. Kingma and J.~Ba, ``Adam: A method for stochastic optimization,'' \emph{arXiv preprint arXiv:1412.6980}, 2014.

\bibitem{marjaninejad2019simple}
A.~Marjaninejad, D.~Urbina-Mel{\'e}ndez, and F.~J. Valero-Cuevas, ``Simple kinematic feedback enhances autonomous learning in bio-inspired tendon-driven systems,'' \emph{arXiv preprint arXiv:1907.04539}, 2019.

\bibitem{shafiee2024effect}
A.~Shafiee, R.~Arabzadeh~Bahri, M.~A. Rafiei, F.~Esmaeilpur~Abianeh, P.~Razmara, K.~Jafarabady, and M.~J. Amini, ``The effect of psychedelics on the level of brain-derived neurotrophic factor: A systematic review and meta-analysis,'' \emph{Journal of Psychopharmacology}, vol.~38, no.~5, pp. 425--431, 2024.

\bibitem{bideh2018security}
P.~N. Bideh, M.~Mahdavi, S.~E. Borujeni, and S.~Arasteh, ``Security analysis of a key based color image watermarking vs. a non-key based technique in telemedicine applications,'' \emph{Multimedia Tools and Applications}, vol.~77, pp. 31\,713--31\,735, 2018.

\bibitem{baraeinejad2022design}
B.~Baraeinejad, M.~F. Shayan, A.~R. Vazifeh, D.~Rashidi, M.~S. Hamedani, H.~Tavolinejad, P.~Gorji, P.~Razmara, K.~Vaziri, D.~Vashaee, \emph{et~al.}, ``Design and implementation of an ultralow-power ecg patch and smart cloud-based platform,'' \emph{IEEE Transactions on Instrumentation and Measurement}, vol.~71, pp. 1--11, 2022.

\bibitem{ames2017hybrid}
A.~D. Ames and I.~Poulakakis, ``Hybrid zero dynamics control of legged robots,'' pp. 292--331, 2017.

\bibitem{vukobratovic2004zero}
M.~Vukobratovi{\'c} and B.~Borovac, ``Zero-moment point—thirty five years of its life,'' \emph{International journal of humanoid robotics}, vol.~1, no.~01, pp. 157--173, 2004.

\bibitem{holmes2006dynamics}
P.~Holmes, R.~J. Full, D.~Koditschek, and J.~Guckenheimer, ``The dynamics of legged locomotion: Models, analyses, and challenges,'' \emph{SIAM review}, vol.~48, no.~2, pp. 207--304, 2006.

\bibitem{kuo2007six}
A.~D. Kuo, ``The six determinants of gait and the inverted pendulum analogy: A dynamic walking perspective,'' \emph{Human movement science}, vol.~26, no.~4, pp. 617--656, 2007.

\bibitem{iqbal2014bifurcations}
S.~Iqbal, X.~Zang, Y.~Zhu, and J.~Zhao, ``Bifurcations and chaos in passive dynamic walking: A review,'' \emph{Robotics and Autonomous Systems}, vol.~62, no.~6, pp. 889--909, 2014.

\bibitem{kuo1999stabilization}
A.~D. Kuo, ``Stabilization of lateral motion in passive dynamic walking,'' \emph{The International journal of robotics research}, vol.~18, no.~9, pp. 917--930, 1999.

\bibitem{harada2006dynamics}
K.~Harada, S.~Kajita, K.~Kaneko, and H.~Hirukawa, ``Dynamics and balance of a humanoid robot during manipulation tasks,'' \emph{IEEE Transactions on Robotics}, vol.~22, no.~3, pp. 568--575, 2006.

\bibitem{raphael2010spinal}
G.~Raphael, G.~A. Tsianos, and G.~E. Loeb, ``Spinal-like regulator facilitates control of a two-degree-of-freedom wrist,'' \emph{Journal of Neuroscience}, vol.~30, no.~28, pp. 9431--9444, 2010.

\bibitem{niyo2023alpha}
G.~Niyo, L.~I. Almofeez, A.~Erwin, and F.~J. Valero-Cuevas, ``An $\alpha$-mn collateral to $\gamma$-mns can mitigate velocity-dependent stretch reflexes during voluntary movement: A computational study,'' \emph{bioRxiv}, pp. 2023--12, 2023.

\end{thebibliography}

\end{document}